# Air Pollution Hotspot Detection and Source Feature Analysis using Cross-domain Urban Data

Yawen Zhang
Department of Computer Science,
University of Colorado Boulder
Boulder, CO
yawen.zhang@colorado.edu

Michael Hannigan
Department of Mechanical Engineering,
Environmental Engineering Program,
University of Colorado Boulder
Boulder, CO
michael.hannigan@colorado.edu

Qin Lv
Department of Computer Science,
University of Colorado Boulder
Boulder, CO
qin.lv@colorado.edu

## ABSTRACT

Air pollution is a major global environmental health threat, in particular for people who live or work near pollution sources. Areas adjacent to pollution sources often have high ambient pollution concentrations, and those areas are commonly referred to as air pollution hotspots. Detecting and characterizing pollution hotspots are of great importance for air quality management, but are challenging due to the high spatial and temporal variability of air pollutants. In this work, we explore the use of mobile sensing data (i.e., air quality sensors installed on vehicles) to detect pollution hotspots. One major challenge with mobile sensing data is uneven sampling, i.e., data collection can vary by both space and time. To address this challenge, we propose a two-step approach to detect hotspots from mobile sensing data, which includes local spike detection and sample-weighted clustering. Essentially, this approach tackles the uneven sampling issue by weighting samples based on their spatial frequency and temporal hit rate, so as to identify robust and persistent hotspots. To contextualize the hotspots and discover potential pollution source characteristics, we explore a variety of cross-domain urban data and extract features from them. As a soft-validation of the extracted features, we build hotspot inference models for cities with and without mobile sensing data. Evaluation results using real-world mobile sensing air quality data as well as cross-domain urban data demonstrate the effectiveness of our approach in detecting and inferring pollution hotspots. Furthermore, the empirical analysis of hotspots and source features yields useful insights regarding neighborhood pollution sources.

## CCS CONCEPTS

• **Information systems** → **Data mining**; **Spatial-temporal systems**; • **Applied computing** → **Environmental sciences**.

## KEYWORDS

Air pollution hotspot detection, pollution source feature analysis, cross-domain urban data

## 1 INTRODUCTION

Air pollution is a major environmental problem and has adverse impact on people's health. For people who live or work near sources of air pollution, such as busy highways, rail yards, marine ports, and industries that emit air pollutants, they are at higher risk of the negative health impact caused by exposure to air pollutants [1]. According to the U.S. Environmental Protection Agency (EPA), more than 45 million people live near roadway in the U.S. [2]. More recent studies show the links between air pollution and coronavirus disease 19 (COVID-19), specifically, a small increase in long-term exposure to air pollutants such as nitrogen dioxide ($NO_2$) and particulate matter (PM) can lead to a large increase in COVID-19 infectivity and mortality rate [22, 30]. For health impact assessment as well as air quality management, it is crucial to understand the spatial patterns of air quality, especially areas with high ambient pollution concentrations, which are commonly referred to as air pollution hotspots [12].

However, due to the high variability of air pollutants [28] in both space and time, it is infeasible to detect air pollution hotspots from observations of conventional sparse monitoring stations [18]. As such, mobile sensing has been explored to observe air quality distributions at higher spatial resolution [3]. For example, Google has equipped its Street View vehicles with air pollutant sensors to monitor street-by-street air quality such as black carbon (BC) and $NO_2$ [13]. This new data modality enables the discovery of air pollution hotspots, in particular the fine-scale localized ones [4]. Since each air pollutant is preferentially emitted by specific sources and can have different atmospheric fate based on chemical reactivity and atmospheric transport, the spatial extent of pollution hotspots differ by pollutant. For instance, the hotspots caused by BC and $NO_x$ are more localized, spanning tens to hundreds of meters [33], while those caused by carbon monoxide (CO) and PM2.5 can persist for much greater distances [34]. Most previous works on spatial hotspot detection are conducted on grids and data with uniform sampling density [21, 26]. Identifying pollution hotspots from mobile sensing data is different in two main aspects: (i) raw observations are trajectory-based: they have variable GPS locations and require further conversion or mapping into fixed spatial data format, such as road segments or grid cells; (ii) uneven spatial and temporal sampling: in real-world operations, vehicles are driving at different speeds on different roads, e.g., highways vs. residential streets, and their spatio-temporal coverage is determined by daily driving assignment, all these result in uneven sampling, i.e., some roads may have more samples or denser temporal coverage than the others. In this study, we aim to explore a general approach to detect air pollution hotspots from unevenly sampled mobile sensing data.

For air pollution hotspots, discovering their potential emission sources is also of great importance. In the past decades, researchers and policy makers have focused on the development of emission inventories, which are essential for policy making in air pollution prevention [25]. However, there are numerous local emission sources unevenly distributed spatially, and those are not well represented in the conventional emission inventories [4], especially some unexpected ones. Figure 1 shows some of the representative air pollution hotspots we have detected from mobile sensing data. We examine aerial imagery to develop hypotheses regarding their possible neighborhood pollution sources, and observe that hotspots can be related to various types of emission sources, such as truck or car businesses, industrial areas, and road intersections. Previously, researchers have tried to manually identify plausible emission sources [4, 10], which is time consuming and does not scale. Fortunately, with advances in crowdsourcing and ubiquitous sensing, a large amount of heterogeneous urban data are collected [32]. The cross-domain urban data, such as Point-of-Interest (POI), land use, vehicle idling, and energy consumption data, provide a good representation of local emission sources thus enable automatic association of air pollution hotspots with their plausible sources.

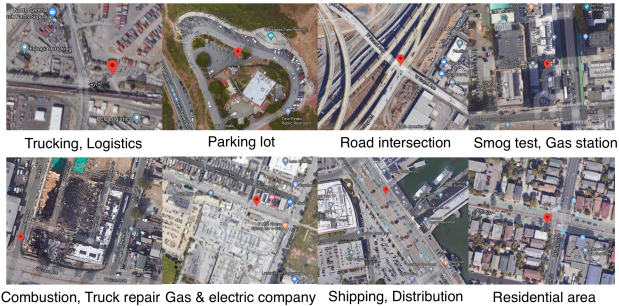

**Figure 1: Representative air pollution hotspots (red markers) detected from mobile sensing data and their plausible neighborhood pollution sources (on Google Maps).**

Motivated by the points above, we propose to leverage mobile sensing and cross-domain urban data to detect air pollution hotspots and characterize local emission sources. In the first phase, we attempt to detect pollution hotspots from mobile sensing data. Essentially, we design a two-step approach for hotspot detection, which aims to address the above-mentioned challenges with mobile sensing. Step 1 involves searching for local pollution spikes on raw observations with a variable-sized sliding window, and step 2 involves identifying pollution hotspots based on the spikes detected in step 1. We utilize sample-weighted clustering method to identify hotspots, while taking into account the uneven sampling issue with mobile sensing data. Our goal is to demonstrate how air pollution hotspots can be identified from mobile sensing data using the proposed approach. This general approach can be applied to different air pollutants and scenarios by tuning the parameters involved. Finding hotspots for specific applications is not the focus of this study. In the second phase, we explore cross-domain urban data to discover possible emission sources. Since ground-truth data on specific emission sources are not available, our goal is to identify plausible pollution sources. Specifically, using the features extracted from heterogeneous urban data, we build models to infer hotspots for cities with and without mobile sensing data. For cities without mobile sensing data, given the fact that most cities have comparable activities, i.e., potentially similar sources of pollution, we investigate the possibility of knowledge transfer. We leverage an adversarial domain adaption technique [11] to learn a domain-invariant feature representation for source and target cities. In the second phase, hotspot inference in different cities provide a soft-validation on the extracted source features. If those features can effectively infer pollution hotspots, it indicates that the related emission sources are well identified and represented.

The contributions of this study are summarized as follows:

- We propose a two-step approach to automatically detect air pollution hotspots from mobile sensing data, which addresses the uneven sampling issue with mobile sensing. The discovery of hotspots can help in narrowing down the regions of interest, which are further investigated by domain experts and regulatory agencies.
- Our analysis of hotspots and source features yields useful insights regarding neighborhood pollution sources. These insights may be directly used to complement conventional pollution inventories.
- As a supplement to empirical analysis, we build hotspot inference models for multiple cities to validate the effectiveness and generalizability of the extracted features in representing local pollution sources.
- We evaluate our approaches using real-world mobile sensing air quality data collected by GSV vehicles in California, and cross-domain urban data from various platforms.

The reminder of the paper is organized as follows. We first summarize the related work in Section 2, and provide preliminaries in Section 3. The overall framework and methodology is presented in Section 4, followed by evaluation experiments in Section 5. Finally, Section 6 provides a discussion on this study, and Section 7 concludes this paper.

## 2 RELATED WORK

**Hotspot Detection.** Hotspot detection aims to discover geographical regions and/or time periods with significantly higher concentration of activities. It plays an essential role in spatial and temporal analysis, and has been conducted in many important applications such as mapping crime rate [16], monitoring disease outbreaks [15], identifying local events [17, 31], and analyzing air quality [4, 10]. The choice of technique for hotspot detection depends on the objective and the type of data, such as grid-based or network data. Different techniques have been proposed for hotspot detection, including local indicators of spatial association [23], regression-based anomaly detection [17], and Kernel Density Estimation (KDE) [16, 20, 31]. KDE is one of the most popular approaches, which identifies hotspots by estimating the underlying probability density function. Zhang et al. [31] define spatial and temporal hotspots as kernel density maxima, and accelerate the KDE-based hotspot detection through pre-computation. Junior et al. [16] propose to detect crime hotspots approximated to the road network with a batch KDE-based algorithm. However, most existing studies are conducted on uniformly sampled data, i.e., data are collected evenly in both space and time. With unevenly sampled data, it is likely that the detected hotspots may have different degree of representativeness due to variations in the number of samples collected. Our study proposes techniques to detect hotspots and tackle the uneven sampling issue with mobile sensing data.

**Pollution Source Discovery.** To analyze pollution sources, atmospheric scientists rely on receptor modeling [5, 27], which leverages measurements of chemical composition data for gas and particle phase air pollutant samples collected at specific sites to quantify source contributions. One commonly used receptor model for source apportionment is the U.S. EPA positive matrix factorization (PMF) [24]. However, due to the measurements required for such modeling, they are usually conducted at a limited number of sampling sites and can only provide information regarding pollution sources at a relatively coarse granularity. This approach can help understand the main pollution sources for a specific region, but is infeasible to discover numerous local pollution sources. There are studies that attempt to discover local pollution sources by exploring air pollution propagation patterns [9, 19]. These approaches can be effective for pollutants that persist for greater distances such as PM2.5, but may not work for localized pollutants such as BC and $NO_x$. Also, the studies do not account for other factors, such as nearby POIs and traffic, thus are unable to provide a comprehensive view of local pollution sources. In more recent studies, researcher have been manually identifying plausible neighborhood pollution sources related to air pollution hotspots by examining related data sources [4, 10]. To automate the process of discovering potential local pollution sources, this study explores the use of cross-domain urban data to study their relationship with pollution hotspots.

## 3 PRELIMINARIES

### 3.1 Definitions

**Definition 1** (Air Quality Dataset). The mobile sensing air quality dataset $\mathcal{A}$ contains a set of points, each denoted by a 5-tuple:

$$\mathcal{A} = \{(c, t, lat, lon, \boldsymbol{a})\},$$

where $c, t, lat, lon, \boldsymbol{a}$ are the unique car ID, timestamp, latitude and longitude (from GPS transmitter), and air pollutants' concentrations. $\boldsymbol{a} = \{a_1, a_2, \ldots, a_q\}$, where $a_i$ represents the observed concentration of the $i$-th air pollutant and $q$ is the total number of air pollutants observed.

**Definition 2** (Car Trajectory). A car's trajectory on a single day is defined as a sequence of GPS points $\boldsymbol{p}_1 \to \boldsymbol{p}_2 \to \ldots \to \boldsymbol{p}_n$, where $\boldsymbol{p}_j = (lat, lon), 1 \leq j \leq n$.

**Definition 3** (City Grids). We partition a city into uniform grids. Each grid $\boldsymbol{g}$ is a square of size $l \times l$ (unit of $l$: $m$). For a city, $\mathcal{G}$ denotes the set of grids within it. City grids are used as basic units for mapping hotspot detection results and matching multi-source data with different spatial characteristics.

**Definition 4** (Air Pollution Hotspots). Air pollution hotspots are locations where multipollutant concentrations exceed nearby ambient levels [4]. In this study, we focus on pollution hotspots which are localized in space (i.e., within a small distance) and persistent in time (i.e., span multiple daytime hours). The air pollution hotspots in a city are denoted by a set of grids $\mathcal{H}_k(\boldsymbol{g})$ from $\mathcal{G}$, where $k$ is the number of grids that are identified as hotspots and $k$ varies across cities.

**Definition 5** (Point-of-Interest Dataset). The POI dataset $\mathcal{P}$ contains a set of POIs, each denoted by a 4-tuple:

$$\mathcal{P} = \{(lat, lon, name, type)\},$$

where $lat, lon$ are the GPS latitude and longitude, and available types of POIs are listed here [1].

**Definition 6** (Vehicle Idling Dataset). The vehicle idling dataset $\mathcal{I}$ contains idling statistics for a city. Each entry is denoted by a 6-tuple:

$$\mathcal{I} = \{(geohash, latmin, latmax, lonmin, lonmax, f)\},$$

where $geohash$ corresponds to a spatial grid of $v \times v$ (unit of $v$: $m$), and $latmin, latmax, lonmin, lonmax$ are the GPS latitudes and longitudes of the grid boundary. $f$ is a set of idling statistics such as cumulative idling time, mean idling time by vehicle type and fuel (i.e., gas or diesel). Detailed description can be found here [2].

### 3.2 Problem Statement

Given the mobile sensing air quality data $\mathcal{A}$ in a city, we aim to detect air pollution hotspots from $\mathcal{A}$ and represent them as $\mathcal{H}(\boldsymbol{g})$. Furthermore, with cross-domain urban data, such as POI ($\mathcal{P}$), vehicle idling ($\mathcal{I}$), land use, and elevation, we aim to extract features from them for hotspot inference in cities with or without mobile sensing data.

## 4 METHODOLOGY

### 4.1 Overview

Figure 2 gives an overview of this study. Our methodology consists of three key components: hotspot detection, source-related feature extraction, and hotspot inference. Firstly, with mobile sensing air quality data, we detect pollution hotspots with a two-step approach. Step 1 involves detecting local spikes from raw observations so as to pick out locations with elevated pollution concentrations. Step 2 aggregates the results of Step 1 by applying sample-weighted clustering to identify robust and persistent hotspots. The sample-weighted clustering technique takes into account the spatial and temporal coverage of each sample. Secondly, with cross-domain urban data, such as POIs, traffic, land use, and elevation, we extract source-related features and create a feature vector representation for each hotspot. Finally, with feature vector representation, we build machine learning (ML) models to infer hotspots in cities with and without mobile sensing data. For cities with mobile sensing data, the hotspot detection results are used as labels for model training. For cities without mobile sensing data, we perform domain adaption between cities for hotspot inference.

### 4.2 Air Pollution Hotspot Detection

Using mobile sensing air quality data, a straightforward and intuitive approach to detect pollution hotspots is to aggregate the available observations into city grids in Definition 3, and then extract hotspots by applying thresholds on air pollutant concentrations. However, our goal is to identify localized and persistent hotspots in Definition 4. The aforementioned approach does not work effectively for this purpose because it lacks two important considerations: (i) *Spatial neighbors*: by comparing with neighboring regions, localized hotspots that are within a limited region can be discovered while the non-local ones can be ignored; and (ii) *Uneven sampling*: as shown in Figure 3, when aggregating all the raw observations into city grids, it is clear that the number of repeated observations and unique daytime hours vary among city grids, which demonstrates the uneven sampling issue with mobile sensing data. To take into account the spatial neighbors and uneven sampling issue, we design a two-step approach to detect pollution hotspots from mobile sensing data: (1) local spike detection, and (2) sample-weighted clustering.

---

[1] https://developers.google.com/maps/documentation/places/web-service/supported_typestable1
[2] https://data.geotab.com/urban-infrastructure/areas-of-idling

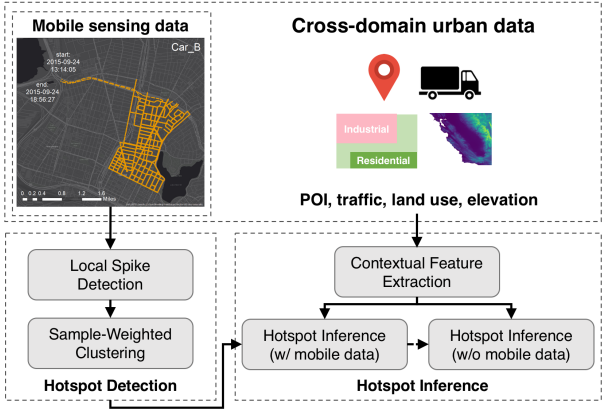

Figure 2: Overview of proposed framework for air pollution hotspot detection and inference.

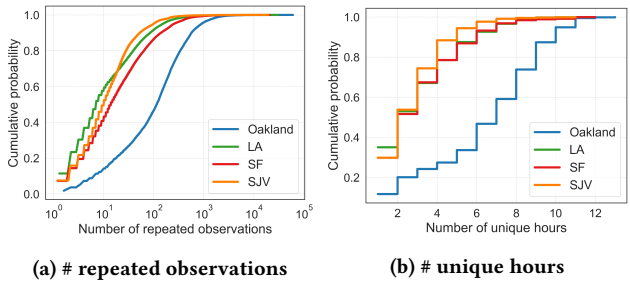

(a) # repeated observations　　(b) # unique hours

Figure 3: Uneven sampling issue with mobile sensing data in different cities: (a) CDF on # repeated observations, and (b) CDF on # unique hours.

**Step 1: Local Spike Detection.** This step aims to extract locations where multipollutant concentrations exceed both the background and neighboring ambient levels. We employ a variable-sized sliding-window based approach to achieve this goal. Specifically, for each day $D$, given mobile sensing data $\mathcal{A}$ of air pollutant $a$ and car trajectory $p_1 \rightarrow p_2 \rightarrow \ldots \rightarrow p_n$, we use a sliding window with varying size to extract segment $M$ of trajectory $p_m \rightarrow p_{m+1} \rightarrow \ldots \rightarrow p_{m+o}$, $1 \leq m \leq n-o$, where $o$ is the number of valid observations in $M$. Essentially, a segment $M$ should satisfy the following constraints:

$$\begin{cases} \textbf{Pollution level:} & a_M \geq a_D, a_M \geq max(ra_L, ra_R) \\ \textbf{Range:} & w_{min} \leq dist(p_m, p_{m+o}) \leq w_{max} \end{cases} \quad (1)$$

In the pollution level constraint, $a_M$ denotes the median air pollutant concentration of segment $M$. $a_D$ and $(a_L, a_R)$ represent the background and neighboring pollution levels, respectively. Here, $a_D$ is computed as the median air pollutant concentration of all observations on day $D$, and $(a_L, a_R)$ are computed as the median air pollutant concentrations of the windows to the left and right of $M$. This constraint requires that $a_M$ not only exceeds the background pollution level $a_D$, but also exceeds at least $r$ of the neighboring pollution level $(a_L, a_R)$. $r$ ($r \geq 1$) denotes the ratio of $a_M$ to its neighbors $(a_L, a_R)$. In the range constraint, $dist(p_m, p_{m+o})$ denotes the distance between the start and end point of $M$. Here, $w$ defines the size range of the sliding window, thus $dist(p_m, p_{m+o})$ should be within the range of $w$, i.e., $[w_{min}, w_{max}]$.

Figure 4 illustrates the process. Given a sliding window of size $w$, we search for local spikes by applying this window on daily car trajectory. For a segment $M$ of trajectory $p_3 \rightarrow p_4 \rightarrow p_5$, $dist(p_3, p_5)$ should be equal to $w$. In terms of pollution level, if $a_M$ exceeds $a_D$, we then compare $a_M$ with $a_L$ and $a_R$. If $a_M$ satisfies both requirements, we obtain a local spike denoted by a 3-tuple $((p_3, p_4, p_5), a_M, t)$, where $t$ is the hour of day. Once a local spike is detected, the search window will move to the end of the right window. The variable-sized sliding-window technique is implemented by repeating the same process with different window sizes from $w$. In this way, local spikes of different ranges can be identified and there is no overlap among the detected spikes.

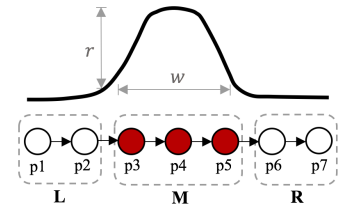

Figure 4: Illustration of local spike detection with variable-sized sliding-window based approach.

For local spike detection, there are three tunable parameters: $a_D$, $r$, and $w$. The first two parameters specify the significance level of a spike, and they are adjustable according to the downstream application. For instance, if we want to identify spikes that are harmful to human health, $a_D$ can be set as the maximum allowable pollutant concentrations established by the U.S. EPA standards [3]. The third parameter $w$ specifies the potential range of a spike, which is largely determined by the air pollutant.

**Step 2: Sample-Weighted Clustering.** Local spikes represent elevated pollutant concentrations, but some of them are not necessarily related to local pollution hotspots. For instance, some spikes are just outliers in observations, such as extremely high pollutant levels caused by encounters with other vehicles' exhaust plumes [4]. It is difficult to avoid such situation since the sensing vehicles are instructed to drive in the normal flow of traffic. Moreover, some of the spikes may be caused by transient pollution sources and only last for a very short time. To exclude these short-lived cases and quantify the robustness of hotspots, we consider spike frequency $S_g$ (i.e., total number of spikes detected in a city grid $g$), since robust hotspots tend to have more spikes than those short-lived ones. Here, we set a minimum threshold of $S_g$ to remove grids with only a few random spikes.

Given the uneven sampling issue with mobile sensing as shown in Figure 3, it is likely that city grids with more repeated observations would have more spikes. For a fair comparison across city grids, besides spike frequency, we include another indicator – spikes' temporal hit rate (THR) – to quantify the persistence of hotspots. Specifically, for a city grid $g$, $THR_g$ is defined as follows:

$$THR_g = \frac{|T_S(g) \cap T_O(g)|}{|T(g)|} \quad (2)$$

where $T(g) = [t_1, t_2, ..., t_z]$. For a specific hour $i$, if there is a valid data point, $t_i$ would be assigned to 1, otherwise 0. Here, $T(g)$ is computed separately for spikes and observations as $T_S(g)$ and $T_O(g)$, and their intersection indicates their temporal consistency. $|T(g)|$ denotes the total number of hours, which is the same for all city grids. For example, if a vehicle typically drives from 8am to 6pm, $|T(g)|$ is set to 10. Higher $THR_g$ indicates higher persistence of the spikes within grid $g$. Compared with $S_g$, $THR_g$ is less impacted by uneven sampling, e.g., there can be grids with a small number of observations but high $THR_g$.

In this step, we firstly aggregate all the spikes into city grids $\mathcal{G}$, and compute $S_g$ and $THR_g$ for each city grid $g$. These two indicators are used to measure the robustness and persistence of $g$. Then, we detect hotspots by adapting the mean shift clustering algorithm [8], which is capable of finding modes of KDE. This procedure is a natural way to identify pollution hotspots, i.e., a pollution source is usually located at a mode of KDE and it causes spikes within its radius-$b$ window. Let $s = (lat, lon)$ be the center of the current window, and $\mathcal{S}(s)$ be the data points inside the window. The mean shift vector would always lead to local density maximum. Here, we incorporate sample weights into the procedure, and the mean shift vector using a Gaussian kernel is computed according to the following equations:

$$m(s) = \frac{\sum_{s_i \in \mathcal{S}(s)} K(s_i - s) s_i}{\sum_{s_i \in \mathcal{S}(s)} K(s_i - s)}, \quad (3)$$

$$K(s_i - s) = W_{s_i} e^{\frac{-||s_i - s||^2}{2b^2}} \quad (4)$$

$$W_{s_i} = S_g THR_g \quad (5)$$

---
[3]https://aqs.epa.gov/aqsweb/documents/codetables/pollutant_standards.html

where $K(\cdot)$ is the kernel function, and sample weight $W_{s_i}$ is computed as the product of $S_g$ and $THR_g$. By assigning weights to grids, the clusters are shifting towards grids that are more robust and persistent. With sample-weighted clustering, some of the isolated grids can be excluded. Finally, we label city grids that form clusters as hotspots, i.e., $\mathcal{H}_k(g)$. In this step, there are two tunable parameters: minimum $S_g$, and window radius $b$. We will provide a detailed parameter study in Section 5.

### 4.3 Source Feature Vector Representation

Air pollution hotspots can be caused by a variety of local pollution sources such as nearby traffic and industry. Finding the exact emission sources that cause the hotspots would require the installation of chemical tracers as well as knowledge from domain experts, which is beyond the scope of this study. Here, using cross-domain urban data, we aim to discover source features that are related to pollution hotspots. As such, we firstly conduct case studies on the detected hotspots so as to select relevant data sources, and then extract features from them to generate source feature vectors for hotspot inference.

*4.3.1 Case Studies.* As shown in Figure 1, we have selected a few representative hotspots detected in the San Francisco Bay Area (SF) for the case studies. Based on our observations, the selected hotspots are related to a few typical categories of pollution sources: (i) car-related businesses (car wash/rental/dealer, smog test, parking lot, and gas station), (ii) truck-involved businesses (logistics, distribution, shipping, and storage), (iii) industrial areas (metal recycling, concrete supply), and (iv) road intersections. While many of these potential pollution sources are directly discernible from map features such as POIs and land use, some hotspots' plausible sources are not easily discernible from the map data, e.g., those within residential areas. They could be caused by unknown business or vehicle related activity such as idling, which is also an important pollution source. Motivated by these observations, we select four different data sources: POIs, vehicle idling, land use, and elevation. We include elevation data because it plays a critical role in the pollutant movement.

*4.3.2 Contextual Feature Extraction.* For each hotspot, we extract the following features and concatenate them to create a source feature vector representation.

- **POI features**. POI features are extracted within a spatial range of 100 m of each hotspot. They are collected using the Google Place API. The POI features include (1) number of POIs in each POI category, (2) number of keywords in POI names (a list of keywords related to pollution sources, which are not included in POI categories), (3) total number of POIs, and (4) POI entropy in terms of category.
- **Land use features**. For each hotspot, we compute its spatial distance to the nearest types of land use, such as industrial, commercial, and residential.
- **Vehicle idling features**. Vehicle idling data is provided by Geotab, which employs mobile sensing to collect various vehicle-related parameters. The vehicle idling features include (1) cumulative, median, and mean idling time, (2) % of different types of vehicles, (3) mean idling time of each vehicle type, and (4) % of different types of engine, such as gas or diesel.
- **Elevation features**. We compute the mean, standard deviation, and concave index (using 3∗3 Laplacian filter) of elevation around each hotspot.

### 4.4 Air Pollution Hotspot Inference

To validate the effectiveness of source features, we build ML models for hotspot inference. Our goal is to test whether or not we can leverage cross-domain urban data to infer pollution hotspots, and how generalizable are those source features, i.e., when inferring pollution hotspots in cities without mobile sensing data.

*4.4.1 City with Mobile Sensing Data.* For a city with mobile sensing data, the hotspot detection results can be used as labels. With source feature vectors, we train binary classifiers for hotspot inference. To deal with data imbalance, we compare two widely used methods, weight balancing and resampling.

*4.4.2 City without Mobile Sensing Data.* Though mobile sensing is being deployed in increasingly more regions to collect fine-grained air quality data, such data is still lacking in many cities. To infer hotspots in cities without mobile sensing data, one way is to directly apply the trained model from cities with mobile sensing data. However, this may not perform as expected due to potential domain shift between cities.

*Domain Shift.* We use the multi-kernel variant of maximum mean discrepancy (MK-MMD) [14] to quantify feature domain differences between cities. A statistical hypothesis test is formulated to reject the null hypothesis $H_0$ that the distributions $\mathcal{F}_s$ and $\mathcal{F}_t$ with $n_s$ and $n_t$ samples respectively are the same. The test statistic is given by:

$$\frac{1}{n_s n_t} \sum_{i=1}^{n_s} \sum_{j=1}^{n_t} k(x_i, x_i) + k(x_j, x_j) - k(x_i, x_j), \quad (6)$$
$$x_i \in \mathcal{F}_s, x_j \in \mathcal{F}_t$$

where $k(\cdot)$ is the kernel function. We use multiple Gaussian kernels with different widths (0.001, 0.01, 0.1, 1) to learn a convex combination of those kernels and then perform the test. Given the test power $\alpha = 0.05$, we compute threshold for the test on the learned kernel. Here, we compare the San Francisco Bay Area (SF) with Los Angeles (LA), and San Joaquin Valley (SJV) which is geographically very different from SF with more farmland and surrounding mountains. As shown in Table 1, for both city pairs, the statistic is significantly higher than the threshold, which proves the feature domain shift between cities. The t-SNE visualization shown in Figure 5 also confirms this. Notably, SJV shows a clear separation from SF in terms of feature distributions.

*Domain Adaption.* Given the differences in feature distributions, we employ domain adaption (DA) methods to learn domain-invariant feature representation. There are DA methods for both shallow and deep ML models. Here, we use the domain adversarial neural network (DANN) architecture to train a model for domain adaption and hotspot inference. This architecture aims to find a balanced feature representation that serves for both domain alignment and primary prediction task, and has achieved competitive results on various tasks [11]. It is suitable for our goal of learning city-invariant feature representation for hotspot inference. The model structure that we implemented is shown in Figure 6, which has three main components: feature extractor, label predictor for hotspot inference, and domain discriminator for city label prediction.

| City Pair | Statistic (Threshold) |
|---|---|
| SF & LA | 6.847 (0.070) |
| SF & SJV | 43.238 (0.074) |

Table 1: MK-MMD results

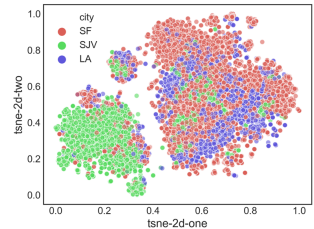

Figure 5: t-SNE visualization

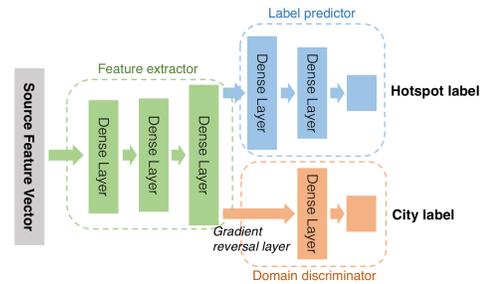

Figure 6: Domain adaption model structure.

## 5 EVALUATION

In this section, we first introduce the experimental setup including datasets and parameter settings. We then present the evaluation results on hotspot detection and inference. We also conduct an empirical analysis on the detected hotspots and source features.

### 5.1 Experimental Setup

*5.1.1 Datasets.* We evaluate our framework in three geographic regions of California: SF, LA, and SJV. We collect mobile sensing air quality dataset and various urban datasets from June 2015 to August 2017, as summarized in Table 2.

**Mobile Sensing Air Quality Data.** We obtain the mobile sensing air quality data from Google, which has equipped several of its Street View vehicles with air pollution sensors. The Google Street View (GSV) air quality dataset in the city of Oakland is collected from June 2015 to May 2016 and has the densest coverage. In the other areas in SF, LA, and SJV, the dataset is collected from May 2016 to August 2017. The vehicles repeatedly collect weekday daytime air pollutant concentrations on the streets at a sampling rate of 1 Hz. To make the observations consistent across regions, we focus on three key health-relevant air pollutants: BC, NO, and $NO_2$.

**Cross-Domain Urban Data.** We partition geographic regions into 50 m × 50 m city grids, and collect multi-source urban data including Google POIs, OpenStreetMap (OSM) land use data, Geotab vehicle idling statistics (153 m × 153 m grids), and SRTM 90-m elevation data. We use city grids as the basis for collecting data at different spatial formats and/or resolutions.

Table 2: Details of the datasets.

|  | Region | | |
|---|---|---|---|
| Data Type | SF | LA | SJV |
| GSV coverage | 85.29 km² | 32.04 km² | 45.23 km² |
| # City grids | 33,530 | 13,475 | 19,875 |
| # GSV observations | 7,897,599 | 1,966,868 | 2,464,331 |
| # GSV driving days | 381 | 105 | 189 |
| # POIs | 430,013 | 111,280 | 78,719 |
| # Vehicle idling grids | 4,797 | 3,738 | 5,638 |
| Mean elevation | 43.56 m | 54.63 m | 24.60 m |

*5.1.2 Parameter Settings.* The major parameters in hotspot detection include: (1) $a_D$: the background air pollution level, (2) $r$: the minimum ratio of pollutant elevations compared to spatial neighbors, (3) $w$: size range of the sliding window, (4) minimum $S_g$: the minimum threshold of spike frequency for clustering, and (5) $b$: the bandwidth for mean shift clustering. By default, we set $a_D$ to the daily median pollution levels, $r = 1.5$, $w = [50, 60, ..., 100]$ with an intercept of 10 m, minimum $S_g = 10$, and $b = 100$ m. With these parameter settings, we aim to identify significant local spikes and aggregate them into hotspots. In Section 5.2, we present a sensitivity analysis of the key parameters in hotspot detection. In hotspot inference, the major parameters are involved in training the DANN models for different regions. By conducting grid search on hyperparameters, we set the learning rate to 0.0001, batch size to 256 for LA and 1024 for SJV. The key parameter $\lambda$ used to balance the losses of label predictor and domain discriminator is set to 0.01 for LA and 0.1 for SJV. The fact that SJV has larger $\lambda$ is consistent with Figure 5 that SJV has larger domain discrepancy.

## 5.2 Hotspot Detection Results

We evaluate the performance of hotspot detection in Oakland and other areas in SF. GSV collected more samples in Oakland than the other regions. We select these two regions to evaluate the performance with different number of samples.

*5.2.1 Evaluation Metrics.* Here, we consider three main performance metrics: mean elevated level of air pollution ($EA$), robustness index ($RI$), and mean temporal hit rate ($THR$). $EA$ measures the ratio of hotspot pollution levels to the background levels, which is defined as follows:

$$EA = \frac{\sum_1^q \frac{a_{g,i}}{a_{D,i}}}{q}, 1 \le i \le q \quad (7)$$

where $a_{g,i}$ is the pollutant concentration at hotspot $g$, and $EA$ computes the mean elevated level of multiple pollutants. Higher $EA$ indicates higher elevated level. To quantify the robustness of hotspots, we randomly separate the raw datasets into two equal-sized parts, and conduct hotspot detection on both of them. $RI$ measures the robustness of the detected hotspots, which is defined as follows:

$$RI = \frac{\frac{|\mathcal{H}_{k_1}(g) \cap \mathcal{H}_{k_2}(g)|}{|\mathcal{H}_{k_1}(g) \cup \mathcal{H}_{k_2}(g)|}}{hr} \quad (8)$$

where $H_{k_1}(g)$ and $H_{k_2}(g)$ are the hotspots detected on the two parts, and $k_1$, $k_2$ are the number of hotspots. We use the Jaccard index to measure their similarity. The higher the Jaccard index, the more similar these two parts, and the more robust the detected hotspots. We compute $RI$ as the ratio of Jaccard index to the hotspot percentage $hr$, i.e., the average percentage of valid city grids that are detected as hotspots. Higher $RI$ indicates higher robustness. Moreover, we use mean $THR$ to measure the persistence of hotspots, and higher $THR$ indicates higher persistence.

*5.2.2 Baseline Methods.* We compare our approach with the following baseline methods:
- **SDO**: the spike detection only (SDO) method only performs spike detection and label spikes as hotspots.
- **TNAS**: the threshold on the number of aggregated spikes (TNAS) performs spike detection first, and then applies a threshold on the spike frequency. Those above the threshold are detected as hotspots.

*5.2.3 Comparison results.* We compare different detection methods in Oakland and other areas in SF in Table 3 and 4. Among all the methods, our approach achieves the best performance on all metrics in both regions, which proves that our approach can detect more robust and persistent hotspots. All the $EA$ values are higher than $r$, which demonstrates the effectiveness of the spike detection process. SDO detects much more hotspots but they are the least robust and persistent, which is consistent with the fact that there are a lot of short-lived spikes. Compared with TNAS, our approach adds $THR$ into sample weights for clustering thus identify more persistent hotspots. We also observe that the mean $THR$ in Oakland is higher than that of other areas in SF, which indicates that the number of samples collected can impact the quality of hotspots.

Table 3: Comparison with baseline methods on hotspot detection (Oakland in SF).

| Method | EA | RI | THR | # hotspots |
|---|---|---|---|---|
| SDO | 2.234 | 1.017 | 0.242 | 6521 |
| TNAS | 2.283 | 2.706 | 0.382 | 2357 |
| Our Approach | 2.303 | 4.540 | 0.424 | 1484 |

Table 4: Comparison with baseline methods on hotspot detection (other areas in SF).

| Method | EA | RI | THR | # hotspots |
|---|---|---|---|---|
| SDO | 2.197 | 1.149 | 0.167 | 15225 |
| TNAS | 2.207 | 4.153 | 0.273 | 3207 |
| Our Approach | 2.312 | 10.376 | 0.321 | 1688 |

### 5.2.4 Parameter Sensitivity Analysis.
We choose two key parameters $b$ and minimum $S_g$ in the clustering process to study their effects on hotspot detection. As shown in Figure 7, compared with $b$, the threshold on $S_g$ has a bigger impact on the number of hotspots detected as well as the mean $THR$. With a higher threshold on $S_g$, the number of hotspots decreases while the $THR$ increases, indicating that the grids with more spikes tend to be more persistent. For $b$, the pattern is the opposite: in general, with a larger $b$, the number of hotspots increases while the mean $THR$ decreases. Setting $b$ as 0.1 km achieves the highest $THR$, which demonstrates the locality characteristic of hotspots related to BC, NO, and NO2.

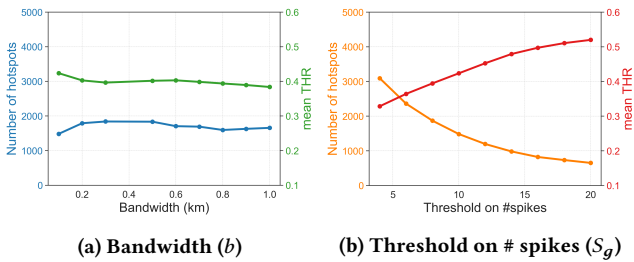

(a) Bandwidth ($b$)    (b) Threshold on # spikes ($S_g$)

Figure 7: Parameter sensitivity analysis.

### 5.2.5 Detection Results.
With our approach, in SF, we detect 3,076 hotspots from 33,530 grids. The percentage of hotspots is about 8.4%. In LA, we detect 620 hotspots from 13,475 grids, and the percentage of hotspots is about 4.4%. In SJV, we detect 239 hotspots from 19,875 grids, which has the lowest hotspot percentage 1.2%.

## 5.3 Empirical Analysis on Hotspots and Source Features

The extracted features are of great importance in capturing potential pollution sources related to hotspots. Figure 8 shows some of the representative features in each category to illustrate their effectiveness in discriminating hotspots and non-hotspots. In Figure 8a, compared with non-hotspots, hotspots have higher number of surrounding POIs in categories such as car repair, gas station, parking, moving company, meal takeaway, etc. In terms of POI name keywords, we observe in Figure 8b that hotspots have higher number of POIs with name keywords such as "trucking", "wheels", "smog", "towing", and "intersection". Both POI categories and name keywords provide a coverage of nearby pollution activity and business. For land use features shown in Figure 8c, compared with non-hotspots, hotspots are closer to industrial, commercial and parking areas. It is a bit surprising that hotspots are not closer to fuel, and this is due to the problem that there is a lot of missing information on fuel in the crowd-sourced OSM data. This observation demonstrates the importance of incorporating multi-source data. For example, the POI information on fuel (e.g., gas station) is of better quality than the OSM fuel data. In terms of vehicle idling time, as shown in Figure 8d, hotspots have much longer cumulative idling time than non-hotspots, which indicates that vehicle idling is very likely a major contributor to some hotspots. In Figure 8f, we observe that hotspots generally have lower elevation than non-hotspots.

This is because locations at lower elevation are more likely to trap air pollutants, resulting in higher pollutant concentrations.

## 5.4 Hotspot Inference Results

With the extracted features, we conduct hotspot inference in two scenarios: (1) City with mobile sensing data: SF, and (2) City without mobile sensing data: LA, SJV. For all three regions, we use the detected hotspots as ground truth for performance evaluation.

### 5.4.1 Evaluation Metrics.
Here, we adopt two metrics to evaluate the performance of hotspot inference: F1 score and Area Under Curve (AUC). AUC is an appropriate metric to evaluate models with imbalanced data [7].

### 5.4.2 Comparison Methods.
For hotspot inference in city with labels, we compare the following strategies to deal with data imbalance (i.e., more non-hotspots than hotspots):

- **Weight balancing (W)**: This method balances the data by altering the weight of each sample, e.g., giving more weight to the minority class when computing the loss.
- **Resampling (R)**: This method combines oversampling (i.e., SMOTE [6]) and undersampling approach, and aims to set an appropriate majority to minority class ratio. To achieve this, we conduct grid search on the oversampling and downsampling ratios.

For hotspot inference in city without labels, we compare the following DA approaches:

- **CORAL**: This method [29] minimizes the domain shift by aligning the second-order statistics of source and target distributions.
- **DANN**: This is a classic adversarial DA method [11] for learning domain-invariant representations.

### 5.4.3 City with Mobile Sensing Data.
For a city with mobile sensing data, the hotspot labels are generated by the detection method. We conduct 5-fold cross-validation on the detection results of SF. Table 5 presents the results of Logistic Regression (LR) and Random Forest (RF) with different strategies for class imbalance. With all the extracted features, RF with the resampling strategy achieves the highest F1 score of 0.682. Compared with F1 score, AUC is less affected by class imbalance. In general, RF performs better than LR in terms of F1 score and AUC, and adding the strategy to deal with class imbalance improves the F1 score in both classification models. Table 6 shows that using all the features outperforms every single feature set. Land use features perform the best, followed by vehicle idling, POI, and elevation features.

Table 5: Comparison of different imbalanced learning strategies ("/" means no strategy applied).

|        | LR       |       | RF       |       |
|--------|----------|-------|----------|-------|
| Method | F1-score | AUC   | F1-score | AUC   |
| /      | 0.208    | 0.863 | 0.595    | 0.966 |
| W      | 0.379    | 0.865 | 0.584    | 0.968 |
| R      | 0.447    | 0.864 | 0.687    | 0.966 |

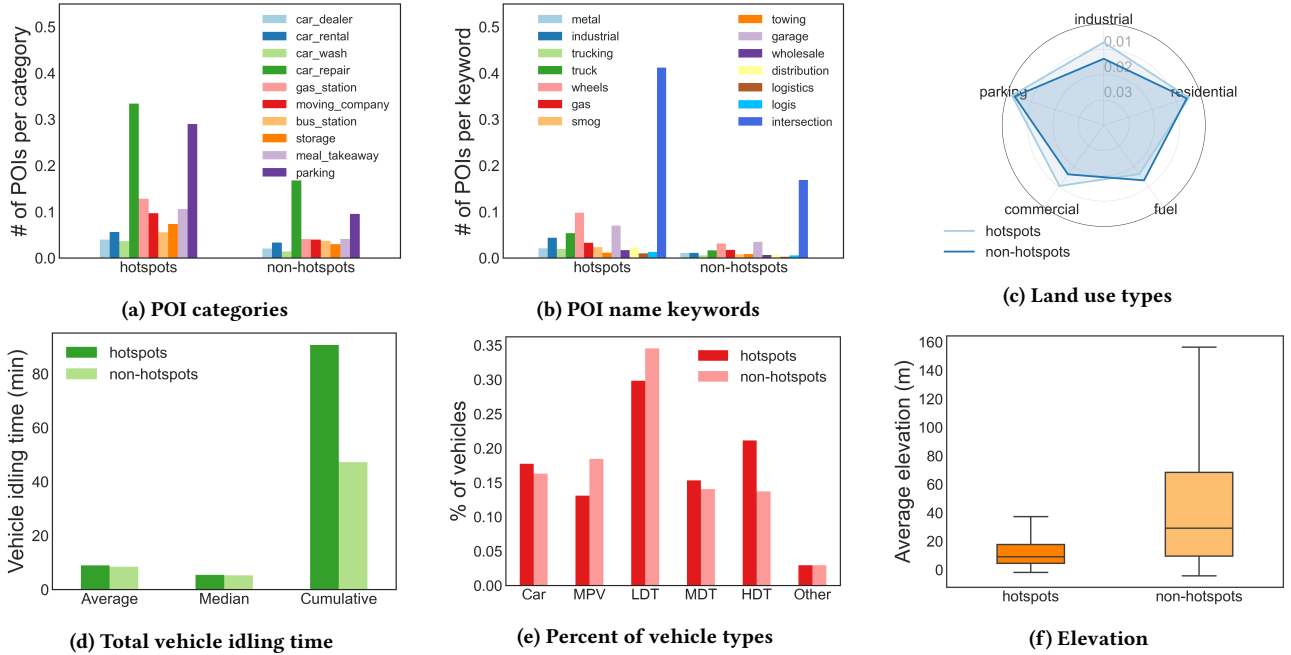

Figure 8: Comparison of hotspots vs. non-hotspots on various features including number of POIs per category, number of POIs per name keyword, distances to different land use types, vehicle idling time (MPV: multi-purpose vehicle, LDT, MDT, and HDT: light, medium, and high duty truck), and elevation.

Table 6: Comparison of different feature sets.

| Feature sets (# features) | F1-score | AUC |
| --- | --- | --- |
| All features (149) | 0.693 | 0.966 |
| Land use features (6) | 0.644 | 0.953 |
| Vehicle idling features (18) | 0.479 | 0.892 |
| POI features (122) | 0.466 | 0.855 |
| Elevation features (3) | 0.304 | 0.795 |

Table 7: Comparison of different DA methods ("/" means no DA method applied).

| | SF → LA | | SF → SJV | |
| --- | --- | --- | --- | --- |
| Method | F1-score | AUC | F1-score | AUC |
| R-/ | 0.194 | 0.753 | 0.060 | 0.773 |
| R-CORAL | 0.195 | 0.762 | 0.060 | 0.781 |
| R-DANN | 0.308 | 0.787 | 0.152 | 0.788 |

5.4.4 *City without Mobile Sensing Data.* For a city without mobile sensing data, there are no labels from hotspot detection. To compare the performance of different DA methods, we use SF as the source city, and LA/SJV as the target city, respectively. Here, we apply the resampling strategy to deal with class imbalance in all models. As shown in Table 7, DANN achieves the best F1 score and AUC, and CORAL only has minor improvement compared with not applying any DA method. The F1 score in LA is higher than that of SJV since SJV has a larger domain shift than LA. Compared with SF, there is a clear performance degradation in LA and SJV, especially for SJV. The recall of the models in these two regions is lower than that of SF, which indicates that some of their hotspots are not well captured in the source city. For example, in SJV, some of its pollution sources come from farming activities like irrigation (with pumps) and wood-burning, which are not included in the source city SF. This observation demonstrates the importance of extending the regions to cover diverse source types.

## 6 DISCUSSIONS

Despite recent advances in large-scale spatial and temporal modeling of air pollutants with ML and deep learning (DL) techniques, there have been few studies on automatic methods to examine the fine-scale patterns of air quality. In this work, we focus on air pollution hotspot detection and source feature analysis. The fine-scale patterns discovered in this work have a lot of real-world use cases, such as air quality management and health impact assessment.

This study provides several promising directions for future research. One direction is to collaborate with public health research teams for hotspot assessment. Those teams can assess the health effects of the detected hotspots in a way that will allow for improved ability to link a hotspot to a concern and then an action. Once the health research team links the elevated pollutant concentration to health effects, then a regulatory agency (and the public) can fix the hotspots that are causing the most harm. This is also a promising downstream application of this work.

Second, our current experiment is conducted on three pollutants in one state. Another direction is to extend the methods to other pollutants and other regions. The pollution source types discovered in this study match well with the 2017 US National Emissions Inventory for BC in California [4], which proves the effectiveness of the hotspot detection method in locating specific pollution sources. However, there may be more uncertainty in source types for other pollutants and other regions. Therefore, it is essential to generalize the methods to more pollutants and regions.

## 7 CONCLUSION

In this study, we design a two-phase framework to detect and infer air pollution hotspots. In the first phase, we propose a two-step approach including local spike detection and sample-weighted clustering to detect hotspots using mobile sensing data. In the second phase, we leverage cross-domain urban data to extract features related to hotspots, and build models for hotspot inference. We evaluate our framework using real-world datasets. The evaluation results demonstrate the effectiveness of our hotspot detection method and the extracted source features.

---

[4]https://www.epa.gov/air-emissions-inventories/2017-national-emissions-inventory-nei-data